\documentclass[11pt]{article}

\usepackage[margin=1in]{geometry}
\usepackage{amsmath,amssymb,amsthm}
\usepackage{booktabs}
\usepackage{multirow}
\usepackage{hyperref}
\usepackage[numbers]{natbib}
\usepackage{enumitem}
\usepackage{xcolor}
\usepackage{array}
\usepackage{colortbl}
\usepackage{tikz}
\usetikzlibrary{arrows.meta,positioning,shapes.geometric,calc,fit,backgrounds,shadows,decorations.pathreplacing}

\definecolor{altblue}{RGB}{30,90,160}
\definecolor{altred}{RGB}{190,45,55}
\definecolor{altgreen}{RGB}{40,130,80}
\definecolor{altorange}{RGB}{220,130,40}
\definecolor{altpurple}{RGB}{120,70,150}
\definecolor{altgray}{RGB}{90,90,95}
\definecolor{altlight}{RGB}{245,245,248}

\hypersetup{colorlinks=true,linkcolor=blue!60!black,citecolor=blue!60!black,urlcolor=blue!60!black}

\title{\textbf{Learning from Disagreement:}\\[4pt]\textbf{Clinician Overrides as Implicit Preference Signals}\\[4pt]\textbf{for Clinical AI in Value-Based Care}}
\author{Prabhjot Singh\thanks{Corresponding author: \texttt{prabhjot@joinaltitude.com}} \and Abhishek Gupta \and Chris Betz \and Abe Flansburg \and Brett Ives \and Sudeep Lama \and Jung Hoon Son}
\date{\textbf{Altitude}}

\begin{document}
\maketitle

\begin{abstract}
We reframe clinician overrides of clinical AI recommendations as implicit preference data---the same signal structure exploited by reinforcement learning from human feedback (RLHF), but richer: the annotator is a domain expert, the alternatives carry real consequences, and downstream outcomes are observable. We present a formal framework extending standard preference learning with three contributions: a five-category override taxonomy mapping override types to distinct model update targets; a preference formulation conditioned on patient state $s$, organizational context $c$, and clinician capability $\kappa$, where $\kappa$ decomposes into execution capability ($\kappa^{\text{exec}}$) and alignment capability ($\kappa^{\text{align}}$); and a dual learning architecture that jointly trains a reward model and a capability model via alternating optimization, preventing a failure mode we term \emph{suppression bias}---the systematic suppression of correct-but-difficult recommendations when clinician capability falls below the execution threshold. We argue that chronic disease management under outcome-based payment contracts produces override data with uniquely favorable properties---longitudinal density, concentrated decision space, outcome labels, and natural capability variation---and that training environments combining longitudinal outcome measurement with aligned financial incentives are a necessary condition for learning a reward model aligned with patient trajectory rather than with encounter economics. This framework emerged from operational work to improve clinician capability in a live value-based care deployment.\looseness=-1
\end{abstract}

\section{Introduction}\label{sec:intro}

Clinical decision support systems are, by most measures, ignored. A systematic review of medication-related CDS alerts found override rates ranging from 46\% to 96\% across studies \citep{poly2020}. This statistic is universally cited as evidence of a problem: alert fatigue, poor specificity, workflow disruption. The dominant interpretation is that overrides represent system failure---a gap between what the CDS recommends and what the clinician does, where the gap itself is the thing to be minimized.

We argue this interpretation is exactly wrong. An override is not noise. It is a signal---one of the richest, most contextually grounded signals available in clinical AI. When a clinician reviews an AI-generated recommendation, weighs it against their knowledge of the patient and the therapeutic goal, and chooses to act differently, they have produced a labeled preference pair: \emph{in this patient state, under these conditions, I prefer action $a'$ over the system's recommendation $r$}. This is precisely the structure exploited by RLHF to align language models with human judgment \citep{christiano2017, ouyang2022}---except the annotator is a domain expert, the alternatives are clinical actions with real consequences, and the downstream outcomes are observable.

\subsection{Motivation: same patient, different environments, different signals}

Consider a 67-year-old patient with Stage C heart failure, managed by a primary care clinician. Guidelines recommend initiation of guideline-directed medical therapy (GDMT), including SGLT2 inhibitors, which reduce HF hospitalization risk by approximately 26\% \citep{mcmurray2019, packer2020, heidenreich2022}. An AI platform surfaces the recommendation: initiate SGLT2i. The clinician overrides it and refers to cardiology instead.

Under fee-for-service (FFS), this override is economically rational: the referral generates billable encounters, the PCP avoids perceived liability, and there is no penalty for the referral-to-therapy delay. Under an outcome-based payment arrangement (\S\ref{sec:properties}) where the organization is rewarded for time-to-outcome rather than encounter volume, the economics invert---the recommendation and the incentive align. A clinician who overrides here is not acting on payment structure but on something else.

That something else is frequently \emph{clinician capability}. Consider three clinicians receiving the same SGLT2i recommendation:

\begin{itemize}[leftmargin=*,itemsep=2pt,topsep=2pt]
\item \textbf{Clinician A}, an experienced internist, reviews the recommendation, confirms renal function and potassium are in range, and accepts it. This is \emph{informed} preference data.
\item \textbf{Clinician B}, a nurse practitioner two years out of training, recognizes SGLT2i is guideline-recommended but has never initiated one independently. He overrides not because he disagrees with the evidence but because he does not trust his own skill to execute it. He refers to cardiology as a risk-avoidance default.
\item \textbf{Clinician C}, similarly early-career, accepts the recommendation---not after critical evaluation but because she accepts most recommendations (95\% acceptance rate). This \emph{looks} like alignment but is the absence of clinical judgment.
\end{itemize}

A naive preference learning system treats these three interactions identically: two acceptances and one rejection. It concludes two-of-three clinicians agree with SGLT2i; Clinician B's override is an outlier to be down-weighted. In reality, only Clinician A's response carries reliable preference information. Clinician B's override is a signal about capability boundaries, not recommendation quality. Clinician C's acceptance is nearly uninformative.

The preference function governing clinical overrides is therefore not merely conditioned on patient state $s$ and contract context $c$. It is also conditioned on clinician capability $\kappa$---the clinician's domain-specific skill for the action being recommended:
\begin{equation}
P(a' \succ r \mid s, c, \kappa)
\end{equation}
where $\kappa$ is not a static attribute of the clinician but varies by clinical domain and evolves over time, potentially through interaction with the decision support system itself.

This introduces what we term the \emph{dual learning problem}. Override data simultaneously trains two models: a \emph{reward model}---what the right recommendation is for a given state and context---and a \emph{capability model}---which clinicians can reliably evaluate which recommendations. Conflating them is dangerous. If 35 of 50 PCPs override SGLT2i initiation due to skill uncertainty, a naive preference model concludes SGLT2i is a poor recommendation and stops recommending it. The correct inference is the opposite: the recommendation is right, but the clinician population needs support to execute it. The override pattern is a training signal for the capability model, not the reward model.

\subsection{Chronic care changes the signal structure}

The CDS override literature is dominated by acute, episodic settings: medication alerts at prescribing, DDI warnings at order entry, sepsis scores in the ICU. In these settings, the override is a one-shot event---a binary accept/reject with thin context, no follow-up, and no capability signal because the interaction does not recur.

Chronic disease management under a subset of VBC that emphasizes goal achievement---outcome-based payoff structures of the kind detailed in \S\ref{sec:properties}---is structurally different. A clinician managing a 200-patient panel across hypertension, HF, diabetes, and CKD generates hundreds of accept/modify/reject decisions per week on the same population, repeatedly, over months and years. Four properties make this override signal qualitatively different: \emph{longitudinal density} (thousands of interactions per clinician per quarter); \emph{outcome observability} (the platform sees what happens after both accepted and overridden recommendations); \emph{contract-conditioned preferences} (the same clinician rationally prefers different actions for the same patient state depending on $c$); and \emph{capability-conditioned signal quality} (heterogeneous clinician populations make $\kappa$ identifiable). We develop these properties formally in Section~\ref{sec:properties}.

\subsection{Contributions}

\begin{enumerate}[leftmargin=*,itemsep=2pt,topsep=2pt]
\item \textbf{Reframing.} The standard framing---overrides as compliance failures to be minimized---is counterproductive. In domains with expert, outcome-consequential annotators, the override is the most valuable data point the system produces.
\item \textbf{Taxonomy.} A five-category taxonomy distinguishing context, judgment, workflow, protocol, and---new to this framework---capability overrides, each mapping to a distinct update target.
\item \textbf{Formal preference framework.} Bradley-Terry extended with patient state $s$, contract context $c$, and clinician capability $\kappa$ conditioning. The modify action provides strictly more information than the binary comparisons used in standard RLHF.
\item \textbf{Dual learning problem.} A formalization showing that override data must jointly train a reward model and a capability model via alternating optimization. We characterize \emph{suppression bias}---the systematic suppression of correct-but-difficult recommendations---as the failure mode of naive approaches, and show the alternating optimization is the structural remedy.
\item \textbf{Architectural blueprint.} The system components required to close the learning loop in a clinical AI platform, producing a compounding data flywheel.
\end{enumerate}

\subsection{Provenance and scope}

This framework did not arise from theoretical synthesis followed by application. It emerged from operational work to improve clinician capability in a live deployment under outcome-based payment contracts, where early efforts to refine recommendation quality from clinician feedback ran into the precise pathologies the framework now formalizes: overrides treated as noise rather than signal; clinician heterogeneity collapsed into average effects; recommendations that gradually disappeared because the population overriding them lacked the capability to execute them; and outcome data attributed naively to actions in ways that would not survive a causal-inference review. The formal structure presented here---the $(s, c, \kappa)$ conditioning, the dual learning loop, the override taxonomy, the failure modes in \S\ref{sec:failures}---was developed to organize and constrain that empirical work. Validation of specific framework predictions is underway and will be reported separately.

The remainder of the paper is organized as follows. Section~\ref{sec:related} positions this work against CDS override analysis, preference-based RL, and RL in healthcare. Section~\ref{sec:framework} presents the formal framework. Section~\ref{sec:properties} characterizes the properties of override data in VBC chronic care. Section~\ref{sec:architecture} describes the system architecture. Section~\ref{sec:implications} discusses implications for clinical AI design and the broader preference learning literature. Section~\ref{sec:limitations} addresses limitations and research directions.

\section{Related Work}\label{sec:related}

Our framework sits at the intersection of three research communities that have operated largely independently: clinical decision support and override analysis, preference-based reinforcement learning, and reinforcement learning applied to healthcare. Each has developed relevant but incomplete tools for the problem we address. Table~\ref{tab:comparison} summarizes the positioning.

\subsection{Clinical decision support and override analysis}

The study of clinician overrides has a two-decade history rooted in medication safety. Override rates for drug safety alerts in CPOE systems range from 49\% to 96\% \citep{vandersijs2006}, with similar 46\%--96\% ranges reported in more recent systematic reviews \citep{poly2020}. Subsequent work has identified workload and repeated exposure as drivers of alert fatigue \citep{ancker2017}, shown that interface design modulates override rates \citep{hussain2019}, and catalogued structured override reasons \citep{wright2019}. Predictive models of override behavior \citep{mccoy2012, straichman2017} and retrospective appropriateness review \citep{poly2020, rehr2018} are designed to suppress or audit overrides, not to learn from them.

This literature treats override as evaluation metric, not training signal, and is overwhelmingly focused on acute, one-shot interactions. The chronic disease management context we address is structurally different: a clinician managing a 200-patient HF panel receives daily prioritized recommendations that recur, evolve with patient state, and produce observable outcomes over weeks to months. The override pattern is not a sparse event log but a dense time series indexed by patient, clinician, domain, and contract. No existing CDS override work treats this as preference data amenable to reward model training.

The closest empirical work is \citet{sivaraman2023}, who identified four behavioral patterns---\emph{ignore}, \emph{rely}, \emph{consider}, and \emph{negotiate}---in ICU clinicians interacting with AI sepsis recommendations. The \emph{negotiate} pattern is a direct empirical observation of the modify action in our framework, but they studied it in a single-encounter experimental setting and did not propose a learning framework. We provide the formal machinery to extract and learn from exactly the negotiation patterns they observed.

\subsection{Preference-based reinforcement learning and RLHF}

\citet{christiano2017} established the two-stage framework of preference learning followed by policy optimization, subsequently applied to LLM alignment by \citet{ouyang2022} and \citet{bai2022}. The standard preference model uses the Bradley-Terry formulation \citep{bradley1952}. Direct Preference Optimization \citep{rafailov2023} eliminates the explicit reward model via reparameterization; Inverse Preference Learning \citep{hejna2023} extends this to sequential settings. \citet{hu2025} addresses regularization artifacts in DPO-style objectives.

This literature has begun to address annotator heterogeneity via multi-annotator reward models and per-annotator noise parameters, but these approaches model heterogeneity as \emph{variation in preference}, not \emph{variation in competence to evaluate}. An annotator who prefers formal over casual tone is providing legitimate preference signal; a novice NP who overrides SGLT2i due to unfamiliarity and an experienced internist who overrides it because potassium is 5.6 are not expressing different preferences---they are providing signals of different \emph{epistemic quality}. This distinction does not exist in current RLHF. Our $(s,c,\kappa)$ formulation is, to our knowledge, the first to disentangle annotator competence from annotator preference in the reward learning loop.

A second gap concerns economic conditioning. In existing RLHF work, preferences depend only on the intrinsic quality of alternatives. In clinical decision-making under VBC, the right action depends on the contract structure under which the clinician operates: a cardiology referral is preferred under FFS and overridden under VBC for the same patient. This conditioning has no analog in the current preference learning literature but is the defining feature of clinical decision-making in the value-based care era.

\subsection{Reinforcement learning in healthcare}

A separate literature applies RL directly to clinical treatment optimization, learning policies from observational data. \citet{komorowski2018} developed the AI Clinician for sepsis using MIMIC-III; subsequent inverse RL work \citep{stith2025} identifies suboptimal clinician decisions by comparing individual behavior to aggregate consensus. This literature positions the AI as teacher: it learns the right policy from data, and clinician deviation from that policy is evidence of clinician error.

Our framework inverts this. We learn from clinician overrides \emph{of} AI recommendations to improve the recommendation system---clinician as teacher, AI as student. The distinction is not merely directional: it reflects a fundamentally different model of the human-AI relationship, and a different goal. The goal is not clinician compliance but recommendation alignment---the point at which overrides decrease because the recommendations have incorporated the judgment that previously motivated them. The work most closely related is \citet{wu2023}, which constrains the RL action space using clinician expertise, but this operates at the policy-constraint level rather than updating the reward model from override signal.

\begin{table}[t]
\centering
\small
\begin{tabular}{lllll}
\toprule
Property & CDS Override & AI Clinician / IRL & Standard RLHF & This Paper \\
\midrule
Treatment of behavior & Eval.\ metric & Policy to outperform & Preference annot.\ & Preference annot.\ \\
Override = ? & System failure & N/A & Pairwise comp.\ & Structured signal \\
Annotator capability? & No & No & No (uniform) & Yes ($\kappa$) \\
Economic context? & No & No & No & Yes ($c$) \\
Observed alternatives? & No & Yes (off-policy) & No (binary) & Yes (modify) \\
Dual learning? & No & No & No & Yes \\
Direction of inference & Alert $\to$ clinician & Clinician $\to$ optimal & Human $\to$ reward & Override $\to$ reward $+\kappa$ \\
\bottomrule
\end{tabular}
\caption{Comparison of approaches to learning from clinician behavior.}
\label{tab:comparison}
\end{table}

\subsection{The gap}

We take the preference learning machinery from RLHF, extend it with capability and contract conditioning ($\kappa$ and $c$), and apply it to the override signal the CDS literature has documented but never formalized as a learning signal. We inherit the longitudinal, outcome-labeled structure from RL-in-healthcare but reverse the direction of inference: we learn from clinician corrections of the AI rather than from clinician behavior as training data for the AI.

\section{The Override as Preference Signal: A Formal Framework}\label{sec:framework}

This section formalizes the argument developed in Section~\ref{sec:intro}. We define the clinical recommendation interaction (\S\ref{sec:interaction}), propose a taxonomy of override types (\S\ref{sec:taxonomy}), derive the preference learning formulation (\S\ref{sec:preference}), characterize the information content of the modify action (\S\ref{sec:modify-info}), and formalize the dual learning problem (\S\ref{sec:dual}).

\subsection{The clinical recommendation interaction}\label{sec:interaction}

We model the interaction between a clinical AI and a clinician as a repeated game over a patient population. At each time step $t$, the system observes patient state $s_t \in \mathcal{S}$ and generates recommendation $r_t \in \mathcal{A}$. The clinician $k$, with capability profile $\kappa_k \in \mathcal{K}$, observes $r_t$ in the context of $s_t$ and contract environment $c \in \mathcal{C}$, then selects:
\[
\delta_t \in \{\text{accept},\ \text{modify}(a'),\ \text{reject}\}
\]
where \emph{accept} executes $r_t$ as recommended ($a_t = r_t$); \emph{modify}($a'$) executes a different but related action; \emph{reject} takes an alternative action that may or may not be directly related.

Each interaction produces a structured record:
\begin{equation}
I_t = (s_t,\ r_t,\ \delta_t,\ a_t,\ k,\ c,\ o_{t+\Delta})
\end{equation}
where $o_{t+\Delta}$ is the observed patient outcome at follow-up horizon $\Delta$ (e.g., BP at 30 days, hospitalization within 90 days, BNP trend at 60 days). The outcome is not available at decision time but is observed retrospectively, and its inclusion is what makes VBC chronic care uniquely suited to outcome-labeled preference learning (\S\ref{sec:properties}).

Over time, a platform with $K$ clinicians, $N$ patients, and $T$ time steps accumulates a dataset $\mathcal{D} = \{I_t\}$ indexed along four dimensions: patient trajectory, clinician identity $k$, clinical domain $d$, and contract context $c$. This multi-indexed structure enables the disaggregation central to our framework.

\subsection{A taxonomy of override types}\label{sec:taxonomy}

Not all overrides carry the same learning signal. We propose five categories based on the source of the discrepancy, each mapping to a distinct model update target, summarized in Table~\ref{tab:taxonomy}.

\begin{table}[h]
\centering
\small
\renewcommand{\arraystretch}{1.5}
\setlength{\tabcolsep}{8pt}
\begin{tabular}{@{}>{\raggedright\arraybackslash}p{2.6cm}>{\raggedright\arraybackslash}p{4.6cm}>{\raggedright\arraybackslash}p{4.3cm}>{\raggedright\arraybackslash}p{3.1cm}@{}}
\toprule
\textbf{Type} & \textbf{Source of discrepancy} & \textbf{Update target} & \textbf{Example} \\
\midrule
\cellcolor{altblue!10}\textbf{I. Context} &
\cellcolor{altblue!5}Clinician has private patient information not in $s_t$ &
\cellcolor{altblue!5}\textcolor{altblue}{\textbf{State representation}} \newline expand $\mathcal{S}$ &
\cellcolor{altblue!5}\footnotesize\itshape Patient reports medication non-adherence not in chart \\
\cellcolor{altred!10}\textbf{II. Judgment} &
\cellcolor{altred!5}Same information as system; genuine clinical disagreement &
\cellcolor{altred!5}\textcolor{altred}{\textbf{Reward model}} $R_\theta$ \newline direct RLHF-analog signal &
\cellcolor{altred!5}\footnotesize\itshape Experienced internist rejects SGLT2i: K$^+$ = 5.6 \\
\cellcolor{altgray!15}\textbf{III. Workflow} &
\cellcolor{altgray!8}Agreement in principle; different action due to time, staffing, or tool constraints &
\cellcolor{altgray!8}\textcolor{altgray}{\textbf{Neither}} \newline filter from preference dataset &
\cellcolor{altgray!8}\footnotesize\itshape No time to counsel in visit; will address next week \\
\cellcolor{altorange!10}\textbf{IV. Protocol} &
\cellcolor{altorange!5}Institutional protocol or standing order supersedes recommendation &
\cellcolor{altorange!5}\textcolor{altorange!85!black}{\textbf{Contract/context}} \newline expand $c$ to encode protocol &
\cellcolor{altorange!5}\footnotesize\itshape Clinic protocol requires cardiology sign-off on GDMT \\
\cellcolor{altpurple!10}\textbf{V. Capability} &
\cellcolor{altpurple!5}Clinician lacks domain-specific skill or confidence to execute &
\cellcolor{altpurple!5}\textcolor{altpurple}{\textbf{Capability model}} $\kappa_\varphi$ \newline may trigger scaffolding &
\cellcolor{altpurple!5}\footnotesize\itshape NP has never initiated SGLT2i; refers to avoid risk \\
\bottomrule
\end{tabular}
\caption{\textbf{The five-category override taxonomy.} Each override type reflects a distinct source of discrepancy between the system's recommendation and the clinician's action, and each maps to a different component of the dual learning architecture. Types I, II, and V are the most informative for model updating; Type III is filtered out as noise; Type IV expands the contract/context representation.}
\label{tab:taxonomy}
\end{table}

Override type is not directly observed---clinicians do not label overrides as capability-driven versus judgment-driven. Type assignment must be inferred from override patterns; we discuss the inference approach in \S\ref{sec:dual}.

\subsection{The preference learning formulation}\label{sec:preference}

We extend the Bradley-Terry model \citep{bradley1952} with three conditioning variables: patient state $s$, contract context $c$, and clinician capability $\kappa$. We treat clinician feedback as informative variation rather than noise to be suppressed; this aligns with recent calls in the broader preference learning literature for treating annotator disagreement as signal \citep{plank2022}.

\textbf{Capability variable: formal definition.} The capability variable $\kappa$ is introduced informally in \S\ref{sec:intro} and operationalized below as a per-clinician, per-domain, time-varying scalar:
\begin{equation}
\kappa_k^d(t) \in [0, 1]
\end{equation}
where $k$ indexes clinicians, $d$ indexes clinical domains (e.g., HF medication management, CKD progression monitoring), and $t$ indexes time. This is the object that enters the preference weighting $\beta(\kappa)$ in Eq.~\ref{eq:beta-linear} below. We later (\S\ref{sec:dual}) propose a conceptual decomposition $\kappa_k^d = (\kappa_k^{\text{exec}}(d),\ \kappa_k^{\text{align}}(d))$ that distinguishes execution capability from alignment capability. We treat this decomposition as a refinement of the scalar definition for interpretation and supervisor-facing analysis; integrating the 2-tuple into the preference weighting (e.g., as $\beta(\kappa^{\text{exec}}, \kappa^{\text{align}})$ with separate temperature contributions) is left to future work. All loss functions in \S\ref{sec:dual} use the scalar $\kappa_k^d(t)$.

\textbf{Latent reward function.} Let $R^*(s, a, c)$ be the true value of taking action $a$ in state $s$ under contract context $c$. Conditioning on $c$ is a deliberate departure from the clinical RL literature, where reward is typically defined independent of payment model. In VBC, the right action incorporates both clinical value and economic context.

\textbf{Preference model.} For recommendation $r$ with observed alternative $a'$:
\begin{align}
P(\delta = \text{accept} \mid s, c, \kappa) &= \sigma\!\left(\beta(\kappa) \cdot \left[R^*(s, r, c) - R^*(s, a_{\text{default}}, c)\right]\right) \\
P(a' \succ r \mid s, c, \kappa) &= \sigma\!\left(\beta(\kappa) \cdot \left[R^*(s, a', c) - R^*(s, r, c)\right]\right)
\end{align}
where $\sigma(\cdot)$ is the logistic function and $\beta(\kappa)$ is a capability-dependent temperature that modulates signal quality.

The innovation is $\beta(\kappa)$. Standard RLHF uses a scalar inverse temperature for noise level. We replace it with a function of clinician capability:
\begin{equation}
\beta(\kappa_k^d) = \beta_0 + \beta_1 \cdot \kappa_k^d
\label{eq:beta-linear}
\end{equation}
where $\kappa_k^d \in [0, 1]$ is clinician $k$'s estimated capability in domain $d$, $\beta_0 > 0$ is a baseline (even low-capability clinicians provide some signal), and $\beta_1 > 0$ controls additional weight for high-capability clinicians. We adopt the linear form for tractability; monotone-bounded alternatives (e.g., $\beta_0 + \beta_1 \cdot \sigma(\kappa)$) are viable and should be compared empirically.

The effect is intuitive: overrides from high-$\kappa$ clinicians drive large reward model updates; overrides from low-$\kappa$ clinicians are down-weighted. This prevents suppression bias (\S\ref{sec:dual}).

\textbf{Contract-conditioned reward.} The parameterized reward model $R_\theta(s, a, c)$ takes encoded patient state, action, and contract context as inputs; its critical feature is that the same action $a$ can have different reward values under different contract contexts $c$.

\subsection{The information advantage of the modify action}\label{sec:modify-info}

Standard RLHF operates on pairwise comparisons: a single bit of ordinal ranking between two system-generated alternatives. The clinical override interaction provides three response types with different information content:

\textbf{Accept.} The clinician endorses $r_t$. Information content depends on $\kappa$.

\textbf{Reject (with alternative $a'$ observed).} A complete preference pair $a' \succ r_t$. More informative than standard RLHF because the preferred alternative was generated by the clinician, not the system---it may lie outside the system's recommendation distribution.

\textbf{Modify.} The richest signal: (1) a preference pair $a' \succ r_t$; (2) a \emph{proximity} signal---$a'$ is close to $r_t$ in action space, indicating the recommendation was approximately correct; and (3) a \emph{gradient direction}---the difference vector $a' - r_t$ points in the direction of improvement. Modify pairs from high-$\kappa$ clinicians are the most informative class of override and should be prioritized accordingly in the $\beta(\kappa)$-weighted training objective.

\subsection{The dual learning problem}\label{sec:dual}

Override data simultaneously trains two models, and these learning problems must be separated to avoid systematic bias.

\textbf{Reward model.} $R_\theta(s, a, c)$ learns the latent clinical-economic value of actions in context:
\begin{equation}
\mathcal{L}_R(\theta) = \sum_{I_t \in \mathcal{D}} \beta(\kappa_{k_t}^{d_t}) \cdot \log P(\delta_t \mid s_t, r_t, a_t, c;\ R_\theta)
\label{eq:LR}
\end{equation}
where $\beta(\kappa_{k_t}^{d_t})$ down-weights low-capability clinician contributions.

\textbf{Capability model.} $\kappa_\varphi(k, d, t)$ learns each clinician's domain-specific competence:
\begin{equation}
\mathcal{L}_\kappa(\varphi) = \sum_{I_t \in \mathcal{D}} \log P(\delta_t \mid s_t, r_t, \kappa_\varphi(k_t, d_t, t);\ R_{\theta^*})
\label{eq:Lkappa}
\end{equation}
where $R_{\theta^*}$ is the current fixed reward model. The capability model asks: given that we know the correct reward, how well does this clinician's behavior correspond to it in this domain?

\textbf{Joint optimization.} The two models are coupled, solved by alternating:
\begin{enumerate}[leftmargin=*,itemsep=1pt,topsep=2pt]
\item \emph{E-step (capability estimation).} Fix $R_\theta$. Estimate $\kappa_\varphi(k, d, t)$ by measuring how well $R_\theta$ predicts the clinician's override behavior in domain $d$. High prediction accuracy implies high $\kappa$; low accuracy implies low $\kappa$.
\item \emph{M-step (reward update).} Fix $\kappa_\varphi$. Update $R_\theta$ using the $\kappa$-weighted preference likelihood $\mathcal{L}_R(\theta)$.
\item \emph{Iterate.} As $R_\theta$ improves, residuals for low-$\kappa$ clinicians shift, updating $\kappa_\varphi$, which changes weighting in the next $R_\theta$ update.
\end{enumerate}
As $R_\theta$ converges to $R^*$, $\kappa_\varphi$ converges to the true capability distribution---residual overrides not explained by the converged reward model are, by definition, driven by non-reward factors (primarily capability).

\textbf{The circularity concern and outcome anchoring.} Eqs.~\ref{eq:LR}--\ref{eq:Lkappa} share the term $P(\delta_t \mid \cdot)$, which makes the dual learning loop structurally coupled: the reward model learns to predict the preferences of clinicians whose preferences were used to estimate the reward model. Without an external anchor, $R_\theta$ could converge to any internally consistent fixed point---including one that systematically reflects clinician collective bias rather than patient benefit. We address this by treating outcome observations $o_{t+\Delta}$ as an anchor \emph{outside} the dual learning loop rather than inside the loss functions. Specifically, the converged $R_\theta$ is validated against outcome-labeled preference pairs (\S\ref{sec:outcome}) on a held-out validation set: a converged model whose preferred actions correlate with worse downstream outcomes is rejected and the alternating optimization is reinitialized with stronger capability priors. This separates the inner-loop preference learning (which uses clinician judgment as the supervision signal) from the outer-loop validation (which uses patient outcomes as the supervision signal), avoiding the contamination that would arise from optimizing a single objective that mixes both. We acknowledge that this design relies on the causal validity of outcome attribution, which is not automatic; the methodology required to make this anchor robust is discussed in \S\ref{sec:outcome}.

\textbf{Suppression bias.} Consider the naive (uncorrected) model with $\beta(\kappa) = \beta_0$ for all clinicians. Suppose true reward favors $a^*$ (e.g., SGLT2i initiation), but 70\% of clinicians override $a^*$ due to low capability ($\kappa \approx 0.2$ for HF management). The naive model receives a majority override signal against $a^*$ and learns $R_\theta(s, a^*, c) < R_\theta(s, a_{\text{referral}}, c)$---it concludes cardiology referral is better. The $\kappa$-corrected model down-weights these low-capability overrides and assigns dominant weight to the 30\% of high-capability clinicians who accept $a^*$, correctly learning $R_\theta(s, a^*, c) > R_\theta(s, a_{\text{referral}}, c)$.

\begin{figure}[h]
\centering
\begin{tikzpicture}[
    every node/.style={font=\small},
    bar/.style={draw=none,rounded corners=1pt},
    label/.style={font=\footnotesize\sffamily},
    panel/.style={font=\small\sffamily\bfseries},
]

\begin{scope}[xshift=0cm]
    \node[panel,text=altgray] at (1.8,5.4) {A.\ Ground truth};
    \node[font=\scriptsize\itshape,text=altgray,align=center] at (1.8,4.95) {Latent reward $R^*(s, a, c)$};
    \draw[->,thick,altgray] (0,0) -- (0,4.2) node[above,font=\footnotesize] {$R^*$};
    \draw[thick,altgray] (0,0) -- (3.8,0);
    \fill[altgreen,bar] (0.5,0) rectangle (1.5,3.3);
    \fill[altorange,bar] (2.2,0) rectangle (3.2,1.9);
    \node[label,align=center] at (1,-0.4) {$a^*$\\\scriptsize(SGLT2i)};
    \node[label,align=center] at (2.7,-0.4) {$a_{\text{referral}}$};
    \node[font=\scriptsize,text=altgreen] at (1,3.55) {\textbf{high}};
    \node[font=\scriptsize,text=altorange] at (2.7,2.15) {\textbf{low}};
\end{scope}

\begin{scope}[xshift=5.5cm]
    \node[panel,text=altred] at (1.8,5.4) {B.\ Naive model};
    \node[font=\scriptsize\itshape,text=altgray,align=center] at (1.8,4.95) {$\beta(\kappa) = \beta_0$ (uniform)};
    \draw[->,thick,altgray] (0,0) -- (0,4.2) node[above,font=\footnotesize] {$R_\theta$};
    \draw[thick,altgray] (0,0) -- (3.8,0);
    \fill[altgreen!40,bar] (0.5,0) rectangle (1.5,1.4);
    \fill[altorange,bar] (2.2,0) rectangle (3.2,3.0);
    \node[label,align=center] at (1,-0.4) {$a^*$\\\scriptsize(SGLT2i)};
    \node[label,align=center] at (2.7,-0.4) {$a_{\text{referral}}$};
    \node[font=\scriptsize,text=altgray] at (1,1.65) {\textbf{suppressed}};
    \node[font=\scriptsize,text=altorange] at (2.7,3.25) {\textbf{elevated}};
    \node[font=\scriptsize\itshape,text=altred] at (1.8,4.0) 
        {\textcolor{altred}{$\blacktriangleleft$ ranking flipped $\blacktriangleright$}};
\end{scope}

\begin{scope}[xshift=11cm]
    \node[panel,text=altgreen] at (1.8,5.4) {C.\ $\kappa$-corrected};
    \node[font=\scriptsize\itshape,text=altgray,align=center] at (1.8,4.95) {$\beta(\kappa) = \beta_0 + \beta_1 \kappa$};
    \draw[->,thick,altgray] (0,0) -- (0,4.2) node[above,font=\footnotesize] {$R_\theta$};
    \draw[thick,altgray] (0,0) -- (3.8,0);
    \fill[altgreen,bar] (0.5,0) rectangle (1.5,3.1);
    \fill[altorange,bar] (2.2,0) rectangle (3.2,2.0);
    \node[label,align=center] at (1,-0.4) {$a^*$\\\scriptsize(SGLT2i)};
    \node[label,align=center] at (2.7,-0.4) {$a_{\text{referral}}$};
    \node[font=\scriptsize,text=altgreen] at (1,3.35) {\textbf{recovered}};
    \node[font=\scriptsize,text=altorange] at (2.7,2.25) {\textbf{correct}};
\end{scope}

\begin{scope}[yshift=-3.8cm]
    \begin{scope}[xshift=0cm]
        \node[font=\footnotesize\sffamily,align=center,text=altgray] at (1.8,1.2)
            {50 clinicians observe};
        \node[font=\footnotesize\sffamily,align=center,text=altgray] at (1.8,0.85)
            {same patient state $s$};
        \node[font=\footnotesize\sffamily\itshape,align=center,text=altgreen] at (1.8,0.35)
            {Truth: initiate $a^*$};
    \end{scope}
    \begin{scope}[xshift=5.5cm]
        \node[font=\footnotesize\sffamily,align=center,text=altgray] at (1.8,1.5)
            {Observed override pattern};
        \foreach \row in {0,1,2,3,4} {
            \foreach \col in {0,1,2,3,4,5,6} {
                \pgfmathtruncatemacro{\idx}{\row*7+\col+1}
                \ifnum\idx<36
                    \fill[altred!70] ({0.1+\col*0.11},{1.15-\row*0.09}) circle (0.036);
                \fi
            }
        }
        \foreach \row in {0,1,2,3,4} {
            \foreach \col in {0,1,2} {
                \fill[altgreen!70] ({2.5+\col*0.11},{1.15-\row*0.09}) circle (0.036);
            }
        }
        \node[font=\scriptsize,text=altred,align=center] at (0.5,0.45) 
            {\textbf{35} low-$\kappa$};
        \node[font=\scriptsize,text=altred,align=center] at (0.5,0.2) 
            {override};
        \node[font=\scriptsize,text=altgreen,align=center] at (2.7,0.45) 
            {\textbf{15} high-$\kappa$};
        \node[font=\scriptsize,text=altgreen,align=center] at (2.7,0.2) 
            {accept};
    \end{scope}
    \begin{scope}[xshift=11cm]
        \node[font=\footnotesize\sffamily,align=center,text=altgray] at (1.8,1.5)
            {Same data, $\kappa$-weighted};
        \foreach \row in {0,1,2,3,4} {
            \foreach \col in {0,1,2,3,4,5,6} {
                \pgfmathtruncatemacro{\idx}{\row*7+\col+1}
                \ifnum\idx<36
                    \fill[altred!18] ({0.1+\col*0.11},{1.15-\row*0.09}) circle (0.036);
                \fi
            }
        }
        \foreach \row in {0,1,2,3,4} {
            \foreach \col in {0,1,2} {
                \fill[altgreen] ({2.5+\col*0.11},{1.15-\row*0.09}) circle (0.065);
            }
        }
        \node[font=\scriptsize,text=altred,align=center] at (0.5,0.45)
            {down-weighted};
        \node[font=\scriptsize,text=altred,align=center] at (0.5,0.2)
            {(low weight)};
        \node[font=\scriptsize,text=altgreen,align=center] at (2.7,0.45)
            {\textbf{dominant}};
        \node[font=\scriptsize,text=altgreen,align=center] at (2.7,0.2)
            {(high weight)};
    \end{scope}
\end{scope}

\begin{scope}[yshift=-5.2cm]
    \node[font=\scriptsize\itshape,align=center,text=altred,text width=4cm] at (7.3,0)
        {Naive model reads majority:\\ $a^*$ learned as suboptimal};
    \node[font=\scriptsize\itshape,align=center,text=altgreen,text width=4cm] at (12.8,0)
        {High-$\kappa$ signal dominates:\\ $a^*$ learned as correct};
\end{scope}

\end{tikzpicture}
\caption{\textbf{Suppression bias and its correction.} Panel A shows the true latent reward $R^*$ favoring action $a^*$ (e.g., PCP-level SGLT2i initiation) over cardiology referral. Panel B: with uniform preference weighting $\beta(\kappa) = \beta_0$, a population of 50 clinicians in which 70\% have low capability for the domain ($\kappa \approx 0.2$) produces a majority override signal against $a^*$. The naive reward model learns the inverted ranking and suppresses the correct recommendation. Panel C: with capability-conditioned weighting $\beta(\kappa) = \beta_0 + \beta_1 \kappa$, overrides from high-capability clinicians dominate the reward update; the model recovers the correct ordering from the same underlying data.}
\label{fig:suppression}
\end{figure}
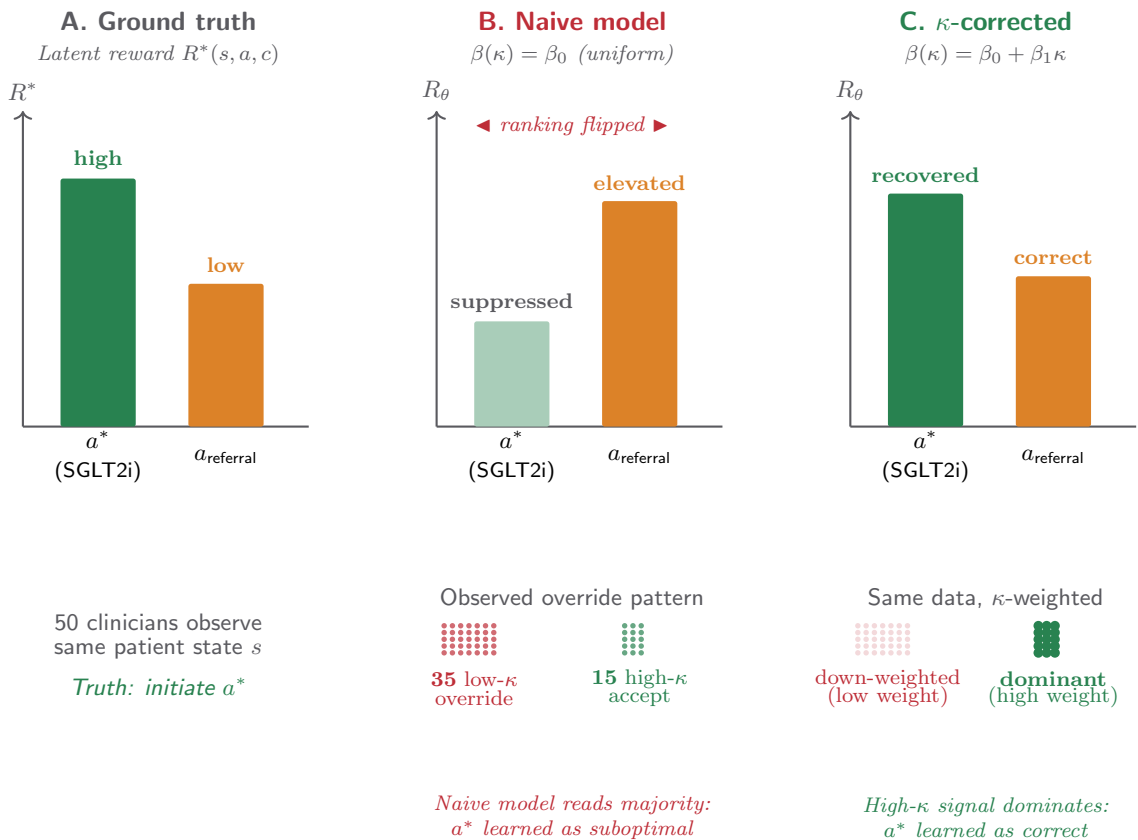

We term this failure mode \emph{suppression bias}: the systematic suppression of correct-but-difficult recommendations in environments where mean clinician capability for the relevant domain is below the execution threshold. Figure~\ref{fig:suppression} illustrates the mechanism. Suppression bias is most dangerous precisely where clinical AI has the most potential value---complex therapeutic initiations that guidelines recommend but many clinicians lack confidence to execute without support.

\textbf{Dynamic capability.} $\kappa_k^d(t)$ is not fixed; it evolves as the clinician gains experience. We decompose:
\begin{equation}
\kappa_k^d = \left(\kappa_k^{\text{exec}}(d),\ \kappa_k^{\text{align}}(d)\right)
\label{eq:kappa-decomp}
\end{equation}
where $\kappa^{\text{exec}}$ is \emph{execution capability} (can the clinician safely perform the action in domain $d$?) and $\kappa^{\text{align}}$ is \emph{alignment capability} (does the clinician's default practice orient toward optimal patient outcomes rather than locally rational FFS behavior?). This creates a second-order learning loop: the system learns which recommendations to make (reward model), which clinicians can execute them (capability model), and which support interventions accelerate clinician capability development.

\textbf{Identifiability.} The dual learning problem is identifiable only when the clinician population has heterogeneous capability. If all clinicians on a platform had identical capability profiles, every override would look the same and the capability model would be unidentifiable. The staffing heterogeneity of VBC chronic care---physicians with $10{+}$ years of chronic disease management, mid- and early-career NPs, care managers with focused scopes---provides this variation organically. On a platform with 50 clinicians managing HF, the override rate for SGLT2i might be 15\% among experienced internists, 55\% among mid-career NPs, and 80\% among early-career NPs. The naive model treats the pooled 55\% as evidence of moderate miscalibration. The dual learning framework decomposes this: the 15\% high-$\kappa$ rate is the reward-relevant signal (the recommendation is correct for $\sim$85\% of firing states); the gradient from 15\% to 80\% as $\kappa$ decreases is the capability-relevant signal. This decomposition is only possible because clinician capability is heterogeneous.

\subsection{Failure modes beyond suppression bias}\label{sec:failures}

Suppression bias is the failure mode the framework is designed to prevent, but it is not the only one. We name four additional failure modes that any deployed implementation must explicitly monitor; the dual learning architecture does not automatically prevent any of them.

\textbf{Amplification bias.} If high-$\kappa$ clinicians share a correlated bias---practice-pattern lock-in, regional training norms, organizational culture---the $\beta(\kappa)$ weighting amplifies that shared bias rather than averaging it away. The framework treats high-capability clinicians as the gold-standard supervision signal; if that signal is itself systematically miscalibrated relative to patient benefit, the reward model converges to the shared miscalibration faster than it would under uniform weighting. Detection requires comparing $R_\theta$ against external benchmarks (guidelines, multi-site comparison, outcome anchors from \S\ref{sec:outcome}) rather than internal consistency alone.

\textbf{Automation bias.} Clinician C in \S\ref{sec:intro} (95\% acceptance rate, no critical evaluation) is the case where uniform acceptance contains nearly zero preference signal but is treated as endorsement by the reward model. The capability model can in principle detect this---a clinician whose acceptances do not correlate with outcome quality has low $\kappa^{\text{align}}$---but only if outcome data has accumulated. In the cold-start period, automation-bias acceptances are weighted equivalently to informed acceptances, which biases early reward updates toward whatever the system currently recommends. The mitigation is conservative initial $\beta$ scaling and explicit detection of clinician-level acceptance entropy.

\textbf{The no-self-correction problem.} A subtle and consequential failure mode arises when the system stops surfacing a recommendation because the reward model has down-weighted it: subsequent overrides cannot occur because the recommendation is no longer presented, and the model has no path to discover it was wrong. Concretely: imagine a VBC organization in which clinicians have been organizationally trained to under-refer to cardiology; their override pattern down-weights cardiology referral recommendations; the system surfaces cardiology referrals less frequently; the small subpopulation of patients who genuinely need referral is not flagged; their adverse outcomes register as undifferentiated decompensations rather than as evidence the recommendation was correctly suppressed. This is a stacking failure---contract incentive shapes clinician behavior, $\kappa$-weighting reinforces that behavior in the reward model, and reduced surfacing eliminates the override data that would correct it. High-precision/low-recall scenarios of this type are particularly dangerous because the reward model is internally consistent, override rates are low, and the capability model shows convergence---all three conventional health-metrics indicate success while the system is failing on a clinically important subset. Detection requires periodic surfacing of low-prior recommendations against held-out validation cases, audit against guideline-based benchmarks, and explicit monitoring of recommendations the system has stopped making rather than only recommendations it currently makes.

\textbf{Degraded outcome observability.} The outcome anchor in \S\ref{sec:dual} assumes outcomes are observable. In practice, attributed patients receive care out-of-network, switch insurers mid-year, or experience outcomes (functional decline, quality-of-life change) that are not captured in claims or EHR data. Outcome observability is highest for hospitalization and mortality, lowest for ambulatory deterioration and patient-reported symptoms. Reward model validation against degraded outcome labels under-detects exactly the failure modes that motivate the framework---slow-rolling clinical decline that VBC contracts are designed to prevent. Production deployment must explicitly characterize outcome capture rates by patient subgroup and adjust the validation procedure for outcome missingness.

These failure modes are not exhaustive; we expect others to emerge with deployment experience. The framework is robust to suppression bias by construction but neutral or vulnerable on the rest. Treating ``the dual learning architecture'' as the full safety story would be a mistake.

\section{Properties of Override Data in VBC Chronic Care}\label{sec:properties}

The framework in Section~\ref{sec:framework} is domain-general, but not all domains produce override data with properties needed for it to yield useful models. This section argues chronic disease management under value-based care contracts produces override data with four properties that make it uniquely well-suited to preference learning---properties absent or attenuated in acute CDS, absent entirely in standard RLHF, and present only partially in RL-in-healthcare. We first clarify the value-based care construct we invoke.

\textbf{Note on the contract construct.} The phrase ``value-based care'' covers a wide range of arrangements in practice, from those primarily focused on diagnosis coding accuracy (HCC capture, RAF optimization) to those that place financial weight on clinical management. The framework's requirements are not satisfied by all of these uniformly. What the framework requires from the contract environment is an \emph{aligned incentive geometry}: a payment structure that rewards earlier, correct, definitive clinical action rather than continuous billable activity, and that produces a clean outcome label tied to the action.

The CMS ACCESS model is the clearest current instance. Its outcome-based payoff structure measures each attributed patient at baseline and again at a threshold for each chronic condition (e.g., HbA1c control, blood pressure control), with payment contingent on whether the threshold was achieved within a defined annual window. Three properties of this geometry matter for the framework. First, the payoff is \emph{discrete}: a patient either reached the threshold or did not, producing a binary ground-truth label per patient-year that is far less susceptible to EHR measurement noise than continuous longitudinal metrics. Second, the payoff \emph{rewards time-to-outcome}: a patient who reaches threshold in month two and remains there has the same payoff as one who reaches threshold in month eleven, but the former requires substantially less clinical effort over the remaining ten months---creating a direct incentive for earlier and more definitive intervention rather than ongoing management activity. Third, the payoff is annual rather than per-encounter, breaking the FFS coupling between clinical effort and revenue.

Conventional VBC contracts---capitated arrangements, shared savings with quality gates, risk-bearing ACO participation---can encode aspects of this geometry and have the potential to satisfy the framework's requirements, but they do so unevenly: many in practice retain coding-driven revenue components, encounter-volume incentives, or quality measures that reward documentation over outcome. Throughout this paper, when we appeal to ``outcome-based payment'' we mean the ACCESS-like payoff geometry as the structurally cleanest case; when we use ``VBC,'' we mean the broader class of contracts that approximate this geometry to varying degrees. What matters for the framework is the incentive geometry, not the specific contract instrument.

\subsection{Longitudinal density}

A clinician managing a 200-patient chronic cardiometabolic panel receives $\sim$2--3 actionable recommendations per patient per week, producing $\sim$400--600 recommendation interactions per clinician per week. Over a six-month window, a single clinician generates $\sim$10{,}000--15{,}000 override interactions, each a structured $(s, r, \delta, a, k, c)$ tuple. For a 50-clinician platform, this is 500{,}000--750{,}000 records in six months---a preference dataset comparable in scale to large RLHF annotation campaigns, generated as a byproduct of clinical workflow rather than through dedicated annotation labor. These figures are illustrative operating estimates derived from observed deployment patterns; empirical interaction density depends on the recommendation system's architecture (rule-based, LLM-based, or hybrid), the platform's surfacing thresholds, and panel composition. We use them to characterize the order of magnitude rather than to make a quantitative empirical claim.

The density is not merely volumetric. The repeated-measures structure---same clinician, same domain, same patient state, multiple time points---is what makes the dual learning problem identifiable. The same clinician's response to the same recommendation type across different patients reveals how patient-specific factors modulate the override. Cross-condition responses (SGLT2i initiation vs.\ beta-blocker titration vs.\ diuretic adjustment within HF management) reveal the structure of their capability profile. Acute CDS lacks this repetition; standard RLHF lacks the patient-state conditioning, contract context, and within-annotator temporal structure.

\subsection{Decision space concentration}

VBC chronic care generates override data concentrated on a dramatically smaller, higher-impact decision space than FFS. The payment model itself acts as a filter.

For a stylized 500-patient chronic disease panel: FFS generates $\sim$8{,}000--15{,}000 discrete clinical decisions per year, many routine refills, documentation-driven coding, or low-signal encounters. VBC concentrates on $\sim$1{,}200--2{,}000 trajectory-changing decisions per year, clustered into five categories: medication titration, escalation thresholds, care gap closures, monitoring protocol changes, and acute event responses. This is $5$--$7\times$ fewer decisions, but the reduction is selective---eliminating low-signal decisions while preserving the high-impact ones. The effective preference density (trajectory-changing decisions per interaction, weighted by signal quality) is approximately an order of magnitude greater in VBC than in FFS.

This concentration compounds. As the reward model improves on the curated decision subspace, override rates on well-calibrated decision types decline, causing remaining overrides to concentrate on genuine blind spots---decisions where clinician private information or contextual judgment adds real value. The preference signal per override grows more informative, accelerating reward model convergence precisely where it is needed most. FFS has no equivalent focusing mechanism: the incentive structure continues to generate high volumes of low-signal decisions regardless of model quality.

\subsection{Outcome observability}\label{sec:outcome}

In standard RLHF, the preference signal is all there is---when an annotator selects A over B, there is no subsequent outcome revealing whether A was objectively better. The RLHF literature has extensively debated whether human preferences are a reliable proxy for true quality, with concerns about annotator bias, Goodhart's law, and reward hacking \citep{casper2023}.

Clinical overrides in VBC have a natural resolution: the outcome $o_{t+\Delta}$ is an objective, externally observable signal that validates or invalidates the preference implied by the override. If a clinician overrides SGLT2i initiation and refers to cardiology, and the patient is hospitalized during the referral-to-therapy delay, the outcome provides direct evidence about the quality of the override. If a different patient receives PCP-initiated SGLT2i and BNP decreases 40\% over 60 days with no adverse events, the outcome validates the acceptance.

This creates \emph{outcome-labeled preference pairs}: preference pairs $(a' \succ r \mid s, c)$ subsequently annotated with outcome observations for both the chosen and (through similar patients) counterfactual actions. For HF: time-to-GDMT, time-to-outcome, hospitalization within 90/180 days, pathway cost, and adverse events are all directly observable. The reward model can be trained on the clinician's preference (via Bradley-Terry likelihood) \emph{and} validated against downstream outcome, providing an external check on calibration. This signal structure is richer than any available in standard RLHF and provides ground truth for evaluating the dual learning problem---if the model has correctly classified an override as Type V (capability-driven) rather than Type II (judgment-driven), outcomes should confirm the override led to worse results.

\textbf{Causal attribution is necessary, not automatic.} The outcome anchoring claim above relies on attributing observed outcomes to specific clinical actions, which is a causal claim that observational data does not automatically support. The hospitalization that follows a cardiology referral may have occurred regardless of therapy initiation timing; the BNP improvement that follows SGLT2i initiation may reflect concurrent diuretic adjustment or natural disease variation. Naive comparisons between accepted and overridden recommendation pathways are subject to confounding by unmeasured patient severity, selection on clinician judgment, and time-varying treatment patterns. The framework therefore requires causal inference methodology as a co-resident discipline rather than an optional refinement: propensity score matching and inverse probability weighting to balance covariates between accepted and overridden pathways \citep{austin2011}, $g$-methods (parametric $g$-formula, marginal structural models) for time-varying confounding common in chronic disease management \citep{robins2000, hernan2020}, and the comparative effectiveness research and pharmacoepidemiology toolkits that have matured around exactly this problem of inferring treatment effects from observational clinical data \citep{stuart2010}. We do not develop these methods here; we name them as the necessary discipline whose absence in the ML engineering pipeline would produce confidently biased reward signal. The outcome anchor in \S\ref{sec:dual} is robust only to the extent that this machinery is applied; clinical AI teams building reward models on outcome-labeled data without causal inference expertise should expect to recover associational patterns that may differ systematically from the causal effects the framework requires.

\subsection{Organizational context and the alignment dimension of capability}\label{sec:context}

The contract context $c$ is a property of the organizational environment, not a per-patient variable that clinicians consciously evaluate at each decision point. A clinician seeing two sequential Stage C HF patients---one attributed to FFS, one to VBC---will almost certainly treat them identically. Clinicians do not switch practice styles patient-by-patient based on insurance type. The contract signal operates through three slower organizational channels:

\textbf{Organizational design.} VBC-native organizations build the infrastructure for PCP-level GDMT initiation: formulary access, titration protocols, monitoring workflows, pharmacist support. FFS-dominant organizations typically have not built this infrastructure---the referral pathway exists because no alternative was ever constructed. Override pattern differences reflect what is operationally possible, not what the clinician consciously chooses in the moment.

\textbf{Clinician development and practice norms.} Clinicians trained in environments with aligned outcome incentives develop different skill profiles. A PCP with five years in such an organization has, through accumulated experience and ambient measurement of time-to-outcome and avoidable utilization, internalized a proactive, longitudinal approach to therapy initiation. They are comfortable initiating GDMT because their environment demanded they learn it. An equivalently trained PCP in an FFS-dominant organization may never have been asked to initiate GDMT independently. This is as much a $\kappa$ effect as a $c$ effect.

This means $c$ and $\kappa$ are not fully independent conditioning variables. The identification strategy for separating their effects relies on cross-organizational comparison rather than within-clinician variation: observing the same recommendation for the same patient state across organizations with different contract structures and different (but overlapping) capability distributions.

\textbf{The alignment dimension.} The coupling of $c$ and $\kappa$ motivates the decomposition in Eq.~\ref{eq:kappa-decomp}. Consider the high-$\kappa^{\text{exec}}$ clinician in an FFS organization---board-certified, experienced, clinically knowledgeable, but refers Stage C HF to cardiology not because they lack the ability to initiate SGLT2i but because their organizational environment has trained them to do so. Their execution capability is high; their alignment capability is low. The reward model's job is not to learn that ``SGLT2i initiation is better under VBC'' (trivially unhelpful) but to learn that PCP-level initiation is the clinically correct action for the state, and that organizational environments that build the infrastructure and clinician alignment to execute it produce better trajectories. An AI system trained on override data from an environment without aligned outcome incentives learns a reward model that is internally consistent with that environment's practice patterns but misaligned with patient outcomes---learning suppression bias by design rather than by accident.

\subsection{The compounding signal}\label{sec:flywheel}

In most preference learning settings, the value of each preference pair is approximately constant---the 10{,}000th pairwise comparison is roughly as informative as the first. In VBC chronic care, informational value increases over time through four compounding effects:

\begin{itemize}[leftmargin=*,itemsep=2pt,topsep=2pt]
\item \emph{Frontier concentration.} As the reward model improves, overrides decrease on well-calibrated states and concentrate on genuine blind spots---higher marginal informational value per override.
\item \emph{Evolving capability.} As clinicians gain experience with the platform's scaffolding, $\kappa$ increases. Overrides at month 6 are more likely to reflect genuine judgment than at month 1, without any change to the learning algorithm.
\item \emph{Accumulating outcome labels.} Outcome labels arrive retrospectively (30/60/90-day lags). The month-12 dataset contains $2\times$ the interactions of month 6 \emph{and} outcome labels for the full month-1-to-6 period, enabling retrospective validation.
\item \emph{Cross-condition transfer.} As $\kappa_k^d$ profiles build across domains, rising capability in one domain (hypertension management) updates priors on related domains (HF management), accelerating convergence.
\end{itemize}
These effects combine into a data flywheel: override data becomes more informative, which accelerates reward and capability model convergence, which further concentrates overrides at the frontier and accelerates clinician development. A system that instruments override capture from day one accumulates a compounding learning advantage---a twelve-month head start is not a twelve-month data gap but a twelve-month compounding gap, because early data's value has been amplified through the flywheel.

\section{System Architecture for Override Intelligence}\label{sec:architecture}

The framework in Sections~\ref{sec:framework}--\ref{sec:properties} defines what should be learned. This section describes the system components required to capture, classify, and learn from override interactions. We present the architecture as a blueprint at a level sufficient for implementation by a clinical AI engineering team; the most consequential design decision is one easily overlooked: instrumentation must be designed into the recommendation interaction from the first patient, not retrofitted.

\subsection{Structured override capture}

The foundation is an instrumented interaction producing the record $I_t$ defined in \S\ref{sec:interaction}. When the platform surfaces a recommendation, the response pathway must branch into structured options: \emph{accept} (records $\delta_t = \text{accept}$, $a_t = r_t$); \emph{modify} (captures both $r_t$ and the modification $a'$ as structured data---for medications, the modified drug, dose, route, frequency); \emph{reject} (captures $\delta_t = \text{reject}$ and, where possible, the alternative $a_t$).

Three design constraints: \emph{minimize capture friction}---in a chronic care platform processing dozens of recommendations per hour, even 10 seconds of structured input creates intolerable friction; rely on one- or two-click structured responses. \emph{Capture the alternative, not just the rejection}---a rejection without observed alternative produces a half-formed preference pair; the architecture must link rejected recommendations to the clinician's subsequent action on the same patient. \emph{Maintain outcome linkage}---the outcome $o_{t+\Delta}$ arrives retrospectively; a durable linkage between $I_t$ and the patient's longitudinal stream must allow outcome observations to be appended as they become available.

\subsection{Override classification}

The classifier assigns each override to a taxonomy category from \S\ref{sec:taxonomy}. Since clinicians do not self-label, classification is inferred from four observable signal sources: the \emph{structure of the alternative} (modifications preserving medication class suggest Type I or II; wholesale pathway substitutions suggest Type IV or V); the clinician's \emph{override history} (consistent domain-specific overrides suggest Type V; case-specific overrides suggest I or II); \emph{cross-clinician comparison} (if 85\% of high-$\kappa$ clinicians accept what this clinician rejects, the rejection is more likely capability-driven); and \emph{structured override reasons} where the capture interface collects them (``patient preference,'' ``not comfortable initiating,'' ``institutional protocol'')---a direct if noisy label.

In practice, overrides rarely map cleanly to a single category. The classifier should output a probability distribution over types $P(\tau_i \mid I_t)$ for $\tau_i \in \{\text{I, II, III, IV, V}\}$, propagating uncertainty into dual model training.

\textbf{The three-way dependence.} Introducing the classifier $\tau_\psi$ creates a three-way coupling: $\tau_\psi$ depends on $R_\theta$ and $\kappa_\varphi$ outputs (cross-clinician comparison signals require both); $R_\theta$ depends on $\tau_\psi$ via the type-weighted preference loss (\S\ref{sec:pairs}); and $\kappa_\varphi$ depends on both. This is materially different from the two-model alternating optimization in \S\ref{sec:dual}; convergence properties of three-way coupled optimization are weaker and not guaranteed by the analysis there. We treat the classifier as a slow-moving outer loop updated less frequently than the inner $R_\theta$/$\kappa_\varphi$ alternation, and we initialize $\tau_\psi$ from structured override reasons and clinician self-report where available, bootstrapping with hand-labeled examples for the cold-start period (\S\ref{sec:training}). Empirical convergence properties of the three-way coupled system are not established here; this is a research direction rather than a settled result.

\subsection{Preference pair construction}\label{sec:pairs}

Classified interactions are assembled into preference pairs: \emph{accept pairs} $(r_t, a_{\text{default}})$---the recommendation preferred over the default absent intervention; \emph{reject pairs} $(a', r_t)$ with $a' \succ r_t$---the most directly analogous to RLHF data; and \emph{modify pairs} $(a', r_t)$ with the proximity and gradient signals from \S\ref{sec:modify-info}.

Each pair is weighted by two factors. \emph{Capability weight} $\beta(\kappa_k^d)$: high-$\kappa$ clinician pairs receive higher weight in the reward objective. \emph{Classification weight}: Type II pairs receive full weight; Type V pairs receive reduced weight for reward model training but full weight for capability model training; Type III pairs receive zero weight in both. As outcome observations $o_{t+\Delta}$ become available, they are appended. Concordance rate by override type---whether the accepted/preferred action led to better outcomes---is a key diagnostic metric.

\subsection{Dual model training}\label{sec:training}

The preference pairs feed into the alternating optimization from \S\ref{sec:dual}.

\textbf{Reward model architecture.} $R_\theta(s, a, c)$ takes as input encoded patient state $\phi(s)$ (structured clinical variables and, where available, unstructured signals from clinical notes via LLM-based extraction), encoded action $\psi(a)$, and encoded organizational context $\gamma(c)$ (payer type, contract structure, quality measure targets, organizational capability indicators). Trained via the $\kappa$-weighted preference likelihood $\mathcal{L}_R(\theta)$. Batch training (periodic retraining) is more stable than online updates and appropriate for early-stage deployment.

\textbf{Capability model architecture.} $\kappa_\varphi(k, d, t)$ outputs a scalar capability estimate per clinician-domain-time triple. Inputs: override history in $d$, outcome history in $d$, and temporal features (time since first interaction, cumulative volume). Updated in the E-step by measuring residuals between $R_{\theta^*}$'s predicted preferences and observed behavior.

\textbf{Training cadence.} Weekly capability updates (as override interactions accumulate), monthly reward updates (as outcome labels become available), with a full joint re-estimation quarterly.

\textbf{Cold start.} A new platform has no override history. During the 4--8 week cold-start, recommendations are generated by a guideline-based prior and overrides are equally weighted. $\kappa$ is initialized from proxy signals---credentials, years of experience, specialty training---as deliberately diffuse priors rapidly updated as interaction data accumulates.

\subsection{The closed loop}

Trained models feed back into the recommendation system, closing the learning loop and producing the flywheel in \S\ref{sec:flywheel}.

\textbf{Reward model $\to$ recommendation engine.} The updated $R_\theta(s, a, c)$ directly improves recommendation ranking: the system evaluates reward for each candidate action and surfaces the highest-reward as the primary recommendation.

\textbf{Capability model $\to$ scaffolding engine.} $\kappa_\varphi(k, d, t)$ triggers adaptive support. For low $\kappa^{\text{exec}}$: contraindication checklists, dose titration protocols, monitoring schedules, real-time lab value checks alongside the recommendation (Clinician B receives the SGLT2i recommendation augmented with: ``eGFR 58---above threshold, K$^+$ 4.2---within range, no current ACEi interaction---safe to initiate. Recommended titration: 10mg daily $\times$ 2 weeks, recheck renal panel at day 14''). For low $\kappa^{\text{align}}$: evidence framing connecting the recommendation to the VBC contract's outcome targets. Scaffolding adapts as $\kappa$ evolves---successful execution with support raises $\kappa$ and decreases scaffolding in subsequent interactions.

\textbf{Capability model $\to$ supervisor dashboard.} An aggregate view of clinician capability by domain: which clinicians have gaps in which domains, whether gaps are narrowing, which scaffolding interventions accelerate development. This transforms override intelligence from a model training signal into a workforce development tool.

\textbf{Override rate as convergence metric.} Rather than tracking a single aggregate rate, track $\kappa$-stratified rates: high-$\kappa$ override rate (reflects reward model calibration---target: decreasing); low-$\kappa$ override rate (reflects capability gaps---target: decreasing with scaffolding); override rate gap (size of the capability gap per domain---target: narrowing). When the high- and low-$\kappa$ rates converge toward the same low value, the system has reached a state where recommendations are well-calibrated and the clinician population can execute them---the dual learning problem is approximately solved for that domain. The rate of convergence quantifies the speed at which the AI platform is simultaneously improving its recommendations and the clinical workforce that acts on them.

\section{Implications}\label{sec:implications}

\textbf{For clinical AI system design.} Clinical AI systems generating recommendations to clinicians should be instrumented for structured override capture from first deployment. The interaction record $I_t$ defined in \S\ref{sec:interaction} is the minimal data structure for a learning system---systems tracking only aggregate acceptance rates, or recording overrides as unstructured free text, are discarding their most valuable training signal. This is a design decision that must be made at the architectural level before the first patient: retrofitting structured capture into a system built with a binary interaction model is substantially harder than designing it in from the start.

The taxonomy in \S\ref{sec:taxonomy} is practical, not just theoretical. Rather than tracking one override rate and interpreting high rates as system failure, teams should decompose by type: a high Type II rate among experienced clinicians signals a genuine recommendation quality problem; a high Type V rate signals a scaffolding gap; a high Type III rate signals an operational feasibility problem. Each requires a different response. The suppression bias failure mode (\S\ref{sec:dual}) is the specific risk to design against: any system that updates from clinician feedback without accounting for capability will, over time, suppress its most valuable recommendations. The dual learning architecture is the structural remedy; a minimum viable mitigation is to weight override signals by a capability proxy (experience, credentials, historical domain performance) and monitor whether recommendation content shifts toward lower-complexity actions over time.

\textbf{For the RLHF research community.} Our framework introduces three concepts that generalize beyond clinical AI. \emph{Capability-conditioned preferences}: the insight that annotator competence $\kappa$ should modulate preference weight applies wherever annotators have heterogeneous expertise. In code generation, a senior engineer's preference between two completions is more informative than a junior developer's; in legal review, an experienced attorney's preference is more informative than a paralegal's. The specific failure mode---suppression bias, where correct-but-difficult outputs are suppressed because low-capability annotators systematically disprefer them---is likely present in existing large-scale RLHF deployments but undiagnosed because annotator capability is not modeled. \emph{Context-conditioned reward functions}: the contract conditioning $c$ generalizes to any setting where the right output depends on the context in which the annotator operates, not just the intrinsic quality of the output---content moderation under different platform policies, translation for different target audiences. \emph{The three-valued response structure}: the accept/modify/reject interaction yields strictly more information than pairwise comparison, with the modify action providing a proximity signal and gradient direction binary comparisons cannot. Developing preference learning algorithms that exploit this structure is an open and, we believe, valuable direction.

\section{Limitations and Research Directions}\label{sec:limitations}

\subsection{Limitations}

\textbf{Generalization from a single deployment.} The empirical work that motivated this framework was conducted on a single platform; generalization across deployment contexts, condition portfolios, and contract structures is open. The framework's structure may not survive contact with deployment data in other settings: taxonomy categories may overlap in practice in ways that resist clean classification, capability-conditioned weighting may interact with classifier uncertainty in ways the current formulation does not anticipate, and the failure modes in \S\ref{sec:failures}---particularly the no-self-correction problem---are by their nature difficult to detect from inside the system that produces them.

\textbf{Partial confounding of $c$ and $\kappa$.} As discussed in \S\ref{sec:context}, organizational context $c$ and clinician capability $\kappa$ are partially coupled---VBC organizations systematically produce different $\kappa$ distributions than FFS organizations. Separating their effects requires a platform operating across organizations with heterogeneous contract structures and overlapping capability distributions. In single-organization deployments, $c$ is effectively constant and the dual learning problem reduces to reward-vs-capability decomposition without contract conditioning.

\textbf{Risk of learning clinician biases.} The framework learns from clinician preferences, which may encode demographic, geographic, or socioeconomic biases producing systematically different override patterns for patients of different backgrounds in ways that are not clinically justified. A high-$\kappa$ clinician's override of a recommendation for a minority patient, if treated as high-confidence preference signal, could train the reward model to recommend differently for minority patients even when the clinical indication is identical. The framework does not currently include a fairness constraint; incorporating demographic parity or equalized opportunity constraints into the reward objective is necessary before deployment at scale.

\textbf{Taxonomy assignment uncertainty.} Classification operates on inferred rather than observed labels. Probabilistic assignment mitigates hard misclassification risk, but in low-data regimes (early deployment, rare domains) the classifier may have high uncertainty, leading to noisy $\kappa$ estimates and degraded updates.

\textbf{Temporal dynamics of $\kappa$.} The framework models $\kappa_k^d(t)$ as smoothly evolving, but real capability development may be discontinuous---focused training programs may produce step changes, burnout or role change may produce step decreases. The current formulation does not explicitly model these, which may produce lagged $\kappa$ estimates that temporarily misweight signal.

\textbf{Contract drift.} VBC payment models evolve: HEDIS measures are added, removed, or redefined; risk adjustment methodology changes; benchmark periods shift. When the contract context $c$ changes meaningfully, the reward model's contract-conditioning becomes stale and the framework requires either retraining on contract-segmented data or an explicit drift-detection mechanism that triggers partial reinitialization without invalidating accumulated override data. Designing such a mechanism---and characterizing how much accumulated data remains valid across contract transitions---is a deployment-critical research direction we do not address here.

\subsection{Research directions}

\textbf{Latent patient state.} The patient state $s$ is treated here as observed. In practice, the system's representation is an approximation that may miss critical variables---Type I context overrides are a direct consequence of state representation incompleteness. A natural extension jointly models patient state as a latent variable with hidden states updated from clinical observations, using context overrides as a signal for state space expansion.

\textbf{Outcome metric formalization.} The outcome observation $o_{t+\Delta}$ is defined generically here. Formalizing \emph{time-to-outcome}---the elapsed time from intervention to target clinical state---as a primary outcome metric is a natural specialization for chronic care platforms: time-to-outcome is the outcome most directly responsive to the override patterns this framework captures.

\textbf{Fairness and bias mitigation.} Beyond the fairness limitation noted above, future work should develop bias detection methods identifying when override patterns vary by patient demographics in clinically unjustified ways, and evaluate whether $\kappa$-weighting inadvertently amplifies or attenuates biases present in high-$\kappa$ clinicians' practice patterns.

\textbf{Generalization beyond healthcare.} The dual learning problem is domain-general. Developing methods to estimate annotator capability from preference data alone (without external capability labels), and evaluating whether $\kappa$-weighted training improves reward model quality in non-clinical domains (code generation, content moderation, legal review), is a promising direction.

\section{Conclusion}

Clinical AI systems generate millions of recommendations. Clinicians override many of them. The field has treated these overrides as failures to be minimized. We have argued that this interpretation discards the most valuable signal clinical AI can access.

A clinician who overrides an AI recommendation produces a labeled preference pair: a structured, contextual, expert judgment about what the right action is for this patient, in this state, under this contract, given this clinician's knowledge and capability. This is precisely the signal RLHF uses to align language models with human values---but richer, because the annotator is a domain expert, the alternatives have real consequences, and downstream outcomes are observable.

We have presented a formal framework: an override taxonomy mapping types to distinct model update targets, a preference learning formulation conditioned on patient state, organizational context, and clinician capability, and a dual learning architecture that simultaneously improves recommendation quality and characterizes the clinician population's capability profile. The framework identifies suppression bias as the specific failure mode naive approaches exhibit and provides the architectural remedy. We have argued that what makes the override signal aligned with patient trajectory is not value-based care in the abstract, but a specific incentive geometry: outcome-based payoff structures that reward earlier, correct, definitive intervention---of which the CMS ACCESS model is the clearest current instance.

Two practical implications follow. First, clinical AI platforms that aim to learn from clinician feedback at scale must build the care infrastructure required to produce structured override data themselves: instrumented recommendation interactions, longitudinal outcome linkage, capability-conditioned analytics. This is a substantial engineering investment, but it is also the only path to the supervision signal the framework requires; the data does not exist as a byproduct of conventional EHR deployment. Second, and equally important, the broader policy environment shapes whether the supervision signal is worth learning from. Clinical AI platforms operating under outcome-based payoff structures accumulate aligned reward signal; platforms operating under FFS or coding-dominant contracts accumulate misaligned signal regardless of how sophisticated their learning machinery is. This means that AI developers, payers, and policymakers share a stake in expanding ACCESS-like incentive geometries: not only because such structures improve clinical outcomes directly, but because they are the operational substrate on which clinical AI can learn to improve outcomes further.

The override signal is generated only through live clinical operation at scale, under the right incentive structure. It is the only training signal in clinical AI that gets better the more you use it.


\end{document}